\documentclass{article}

\newcommand{\marlo}{MARL\"{O}}

\usepackage[preprint]{nips_2018}

\usepackage[utf8]{inputenc} % allow utf-8 input
\usepackage[T1]{fontenc}    % use 8-bit T1 fonts
\usepackage{hyperref}       % hyperlinks
\usepackage{url}            % simple URL typesetting
\usepackage{booktabs}       % professional-quality tables
\usepackage{amsfonts}       % blackboard math symbols
\usepackage{nicefrac}       % compact symbols for 1/2, etc.
\usepackage{microtype}      % microtypography

\usepackage{geometry}
\usepackage{xcolor}
\usepackage{todonotes}
\usepackage{amssymb, soul}
\usepackage{hyperref}
\usepackage{listings}
\usepackage[noend]{algpseudocode}
\usepackage{subfigure}

\title{The Multi-Agent Reinforcement Learning in Malm\"{O} (MARL\"{O}) Competition}

\author{
  Diego Perez-Liebana \and Katja Hofmann \and Sharada Prasanna Mohanty \and Noboru Kuno \and Andre Kramer \and Sam Devlin \and Raluca D. Gaina \and Daniel Ionita \thanks{Diego Perez-Liebana, Raluca D. Gaina and Daniel Ionita are with the Game AI group, Queen Mary University of London (UK). Katja Hofmann, Noboru Kuno, Andre Kramer and Sam Devlin are with Microsoft Research (UK). Sharada Prasanna Mohanty is with \'Ecole Polytechnique F\'ed\'erale de Lausanne (Switzerland).}
}

\begin{document}
%\nipsfinalcopy is no longer used

\maketitle

\begin{abstract}
Learning in multi-agent scenarios is a fruitful research direction, but current approaches still show scalability problems in multiple games with general reward settings and different opponent types. The Multi-Agent Reinforcement Learning in Malm\"{O} (\marlo) competition is a new challenge that proposes research in this domain using multiple 3D games. The goal of this contest is to foster research in general agents that can learn across different games and opponent types, proposing a challenge as a milestone in the direction of Artificial General Intelligence.
\end{abstract}

\section{Introduction}

Learning in multi-agent settings is one of the fundamental problems in AI research. Independently learning agents can result in non-stationarity, and the presence of adversarial agents can hamper exploration and consequently the learning progress. Multi-agent settings can be approached as Stochastic $N$-player Games (SGs) \cite{Shapley1953}, where each player interacts with the game environment by observing state observations, sending actions that in turn affect the state of the environment (and other players) and receiving rewards. Reinforcement Learning (RL) is a common approach to learning in SGs~\cite{tan1993multi,littman1994markov} and promises solutions that could be general and applicable to learning in any game.

RL for multi-agent settings has a long and fruitful research tradition \cite{stone2000multiagent,busoniu2006multi}. The goal of a reinforcement learner is formally to maximize its long-term cumulative reward. % while interacting with the environments and the players within it. 
%Depending on the structure of the reward function, 
Games can be competitive (e.g., zero-sum games where one player's reward is the inverse of its opponent's), collaborative (all rewards are shared), or general-reward. The latter are the most realistic for many real-world applications but also notoriously challenging. Even in more restricted purely competitive and collaborative settings, the challenges of learning in the presence of other agents are far from solved. Current solutions only scale empirically in tasks restricted to relatively small environments or with simplifying assumptions.

%Despite the complex challenges that underlie RL in multi-agent settings, 
Recent research progress in multi-agent RL have shown rapid progress in tackling some key challenges \cite{foerster2017stabilising,foerster2017counterfactual,Lowe2017,Sunehag2017}, suggesting that increased research efforts could lead to further breakthroughs. Genralization beyond individual tasks and opponent types is an area with high need and potential for further research. In single agent RL, there is a clear risk to overfitting to individual tasks and specific opponent types. This problem is gradually being addressed in single-agent tasks, but most current empirical work in multi-agent RL is focused on few individual tasks with a single learning agent.% We see one cause of this in the high cost of creating multi-agent tasks and in creating multiple types of learning agents. 

\section{MARL\"{O}: Multi-Agent Reinforcement Learning in Malm\"{O}}

The Multi-Agent Reinforcement Learning in Malm\"{O} (MARL\"{O}) competition is a new challenge that proposes research on multi-agent RL using multiple games. Participants would create learning agents that will be able to play multiple 3D games within Minecraft as defined in the Malm\"{O} platform. The MARL\"{O} competition will run for the first time in 2018 and will feature $3$ games\footnote{https://www.crowdai.org/challenges/marlo-2018}:\\

\vspace{-0.35cm}
\noindent \textbf{Mob Chase:} A collaborative game in which two or more agents and a mob wander a small meadow for a limited amount of time. The agents can either catch the mob - cornering it, leaving no escape path available - or leave the pen through one of the exits. Capturing the mob gives players a high reward ($1$ point) while exiting provides a smaller one ($0.2$). This game is inspired by the variant of the \textit{stag hunt} presented in~\cite{Yoshida08theory}.\\ %Stag hunt\footnote{https://en.wikipedia.org/wiki/Stag\_hunt} is a classical game theoretic game formulation that captures conflicts between collaboration and individual safety.\\

\vspace{-0.35cm}
\noindent \textbf{Build Battle:} A competitive and collaborative game where two teams of agents compete to build a given cuboid structure within a time limit. Agents receive $0.2$ points for correctly placing a block or removing an incorrectly placed block, and $-0.2$ points for incorrectly placing a block or removing a correctly placed block.\\ %There is a time limit proportional to the number of blocks missing from the structure to be built by players.\\

\vspace{-0.35cm}
\noindent \textbf{Treasure Hunt:} A competitive and collaborative game played in an underground dungeon. Each team is formed by \textit{collectors} (who can pick treasures) and \textit{fighters} (who fight foes). The goal is to retrieve a treasure while surviving the enemy entities. All agents on a winning team receive $0.25$ points if their collector gets the treasure, and $0.5$ points if the collector reaches the exit (losing teams receive the negation of these points). If anybody in the team dies, all agents on the team receive $-1$ points. The game ends when the collector player reaches the exit, when a player dies or when the time runs out.  

All games are parameterizable, providing a task space in which potentially endless variants of each domain can be created. Examples of these parameters are weather, block types, number and position of entities, and size of the playing area. Participants can train their agents in any of the possible instances of each game (known as \textit{tasks}). The final evaluation will be performed in particular customizations designed by the organizers and not revealed before the competition deadline.

The participants will be provided an extensive starter kit\footnote{https://github.com/crowdAI/marlo-single-agent-starter-kit/}, with all necessary instructions to download the framework, develop and execute their agents locally for testing in the games included in the benchmark. The starter kit will also include a set of simple tasks and the code for default challenge agents (to compete against, but also as examples for participants to develop their own entries). This approach has been proven successful in previous competitions~\cite{perez2016general,learnToRun,malmoCollAI}, helping entrants to quickly get started with the challenge and easily continue to make iterative changes to their approach.

%The organising team will ensure that the participants are able to quickly get started with the challenge with a few lines of code, and then easily continue to make iterative changes to their approach.

%Agent code needs to be packaged in a docker container before it is submitted for evaluation, and needs to be able to run reliably within a given set of resource constraints. The participants will have the option to either submit their submissions as docker image dumps, or as a source code repository (which can be deterministically converted to a docker image by using binder \textit{https://mybinder.org}). A template repository will be provided in the starter kit which has the binder configuration details for all popular software environments like PyTorch, Tensorflow, chainer-rl, etc; and the participants will also be provided the binder tools ecosystem to help them locally test the conversion of their submissions. This approach ensures that all the submissions are also meaningfully reproducible. 

The final rankings of the competition are computed by means of a play-off tournament. Each stage in the tournament features the $3$ games mentioned here, at least $1$ task for each one of them and $N$ teams playing a round-robin league. The agents must therefore show proficiency in multiple games to move to the next round of the play-off. Each league will have its own ranking table, in which entries are sorted by the sum of scores obtained in all tasks. The top entries of each group progress to the next stage, until reaching the final league, which determines the competition winner.

\section{Conclusions}

The MARL\"{O} competition aims at encouraging research in multi-agent reinforcement learning. In particular, it proposes a challenge in which researchers must attempt approaches that generalize well across games, tasks and opponent types. Not only submitted agents are tested in different games, but these games are also highly parameterizable, with the final configuration of tasks not known to participants beforehand. Additionally, all entries play against multiple agents in the tournament, requiring the agents not to overfit to their opponents.

Research is therefore encouraged and supported by this competition in multiple ways: it makes a set of multi-agent learning tasks publicly available, with a low barrier of entry, but enough room for increasing task difficulty as progress is made. It also creates shared baselines and an evaluation setup that eases comparison between approaches. Finally, it increases awareness of the multi-agent RL challenges and provides a platform for testing and sharing the progress in the field. 

We believe that this is the right way at the right time to invigorate the research community and drive progress in this exciting and important area. The future of the competition will bring different games with new dynamics, richer interactions and further challenges for multi-agent RL reserach. 

\newpage
\bibliographystyle{plain}

\bibliography{bib}

\end{document}